\documentclass{article}

\usepackage[preprint]{spconfa4}
\usepackage{amssymb,amsmath}
\usepackage{framed,multirow}
\usepackage{latexsym}
\usepackage{array}
\usepackage{balance}
\usepackage{booktabs}
\usepackage{caption}
\usepackage{color}
\usepackage{graphicx}
\usepackage{multirow}
\usepackage{times}
\usepackage{url}
\usepackage{threeparttable}
\usepackage{float}

\makeatletter
\newif\if@restonecol
\makeatother

\usepackage[linesnumbered,ruled,vlined]{algorithm2e}
\usepackage{algpseudocode}
\usepackage{amsmath}

\newcommand\etal{\emph{et~al.}}
\newcommand\ie{\emph{i.e.}}

\begin{document}

\title{Document Layout Analysis with Aesthetic-Guided Image Augmentation}

\name{Tianlong Ma$^1$, Xingjiao Wu$^2$, Xin Li$^2$, Xiangcheng Du$^{2,3}$, Zhao Zhou$^{1,3}$,  Liang Xue$^4$,Cheng Jin$^1$}
\address{
    $^1$Fudan University, Shanghai, China~~~~~
    $^2$East China Normal University, Shanghai, China\\
    $^3$Videt Lab, Shanghai, China ~~~~~
    $^4$Zhongshan Hospital, Fudan University, China
}

\maketitle

\begin{abstract}
 Document layout analysis (DLA) plays an important role in information extraction and document understanding.
 At present, document layout analysis has reached a milestone achievement, however, document layout analysis of non-Manhattan is still a challenge.
 In this paper, we propose an image layer modeling method to tackle this challenge.
 To measure the proposed image layer modeling method, we propose a manually-labeled non-Manhattan layout fine-grained segmentation dataset named FPD. As far as we know, FPD is the first manually-labeled non-Manhattan layout fine-grained segmentation dataset.
 To effectively extract fine-grained features of documents, we propose an edge embedding network named  $L\text{-}E^3Net$.
 Experimental results prove that our proposed image layer modeling method can better deal with the fine-grained segmented document of the non-Manhattan layout.

\end{abstract}


\section{Introduction}
\label{sec:intro}

Document layout analysis (DLA) task usually uses semantic segmentation technology to divide images, tables, text, and background in the document layout into different areas, and DLA is a pixel-level classification. DLA plays an important role in document content understanding and  Information extraction, such as optical character recognition (OCR), biomedical event extraction, handwriting recognition, and knowledge extraction~\cite{binmakhashen2019document}.

At present, DLA has reached a milestone achievement, especially for the processing of Manhattan layout documents (\ie~academic documents)~\cite{degtyarenko2021hierarchical, li2021few, wu2021document}.
However, with the improvement of people aesthetics, document design is no longer limited to simple Manhattan layouts~\cite{arroyo2021variational}, to pursue aesthetics, designers began to use more non-Manhattan layouts (\ie~magazine pages)~\cite{sun2005page}.
In addition, the processing of non-Manhattan document layout is the main bottleneck restricting the development of DLA~\cite{shen2020large, lombardi2020deep, bhowmik2021binyas}.

Previously, researchers focused on non-Manhattan layout document layouts, and ignored the processing of non-Manhattan layout document layouts~\cite{bhowmik2021binyas}.
Therefore, the current large-scale datasets used for network training belong to the way of Manhattan document layout (\ie~Publaynet~\cite{zhong2019publaynet}, DocBank~\cite{li2020docbank}, CS-150~\cite{clark2016pdffigures}).
Large-scale datasets dedicated to non-Manhattan document layout research have not yet been proposed.

\begin{figure}[t]
  \centering
  \includegraphics[width=0.92\linewidth]{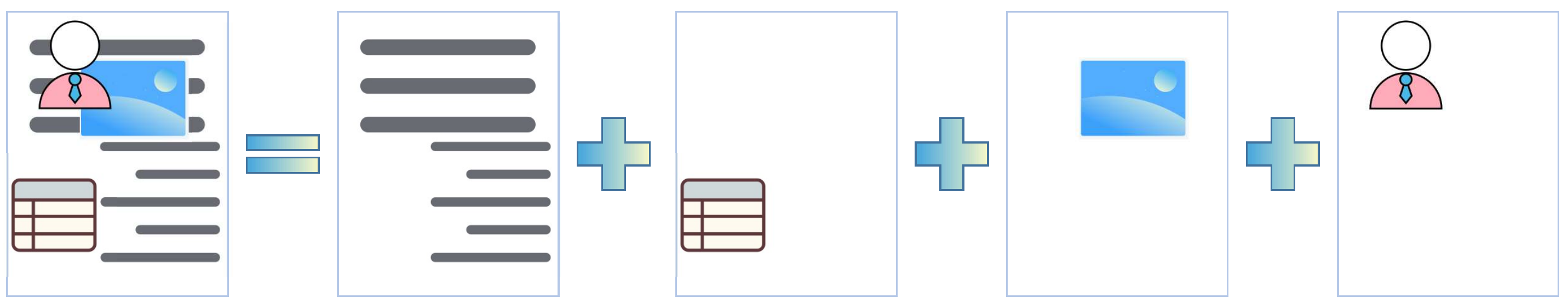}
  \caption{The example of image layer modeling. }
  \label{fig:motivation}
  \vspace{-10px}
\end{figure}

The central issue addressed in this paper is the following:
\emph{How do we construct a non-Manhattan layout documents handle method and conduct evaluate this method effectively?}

Manual labeling of data is labor-intensive, and labeling throughput is limited~\cite{zhang2008active}.
Especially for the non-Manhattan layout document sample.
Most of the current work uses synthetic data instead of manual annotation (\ie~use LATEX as the engine~\cite{yang2017learning}, LayoutGAN~\cite{li2020layoutgan}, layoutVAE~\cite{jyothi2019layoutvae}).
Following this work, we also hope to use a novel model that does not require manual intervention to generate new examples.
In contrast to previous works that use engines for synthetic data, our framework employs the image layer modeling~\cite{chen2020flexible}, we show examples in Fig.~\ref{fig:motivation}.
Compared with the previous method of using engine synthesis, use the image layer modeling has stronger flexibility and robustness.
In addition, the image layer modeling also strictly follows aesthetic standards to be able to produce more natural samples.
In this paper, we propose a method based on layer modeling, which is built upon the Aesthetic-Guided~\cite{deng2017image} and copy-paste~\cite{salehinejad2018image}.
In addition, the non-Manhattan layout datasets are all marked by rectangular boxes, and the results obtained by such marking are inaccurate.
So we re-annotated a new dataset using point-based annotation to evaluate the processing effect of non-Manhattan layout documents.

To summarize, the highlights of this paper are:
(1) We propose a novel  image layer modeling method for the DLA model by incorporating image layer modeling and aesthetic-guided;
(2) We propose a new fine-grained page semantic dataset using point annotation to evaluate the analysis effect of non-Manhattan layout documents;
(3) We propose a lightweight edge enhancement network to better process non-Manhattan layout documents;
(4) Extensive evaluation on two DLA datasets with very competitive results of the proposed image layer modeling approach.

\section{Related Work}
\label{sec:related}
\noindent\textbf{Document layout analysis.}
DLA tasks can be divided into traditional methods and methods based on deep learning~\cite{binmakhashen2019document}.
The traditional method uses the image processing to analyze the document layout(\ie Voronoi-Based Analysis~\cite{kise1998segmentation}, Delaunay Triangulation Analysis~\cite{eskenazi2015delaunay},  Run Length Smearing Algorithm~\cite{wahl1982block}, and DLA Projection Profile~\cite{shafait2010effect}).

With the development of deep learning, more and more deep learning-based DLA algorithms have been proposed, and these algorithms have achieved milestone results due to their superior characteristics.
The primary challenge of deep learning methods is that deep learning models need to be driven by many data.
Many researchers have explored different aspects to obtain a large amount of data.
Zhong~\etal~\cite{zhong2019publaynet}~develop the PubLayNet dataset for DLA by automatically matching the XML representations.
Li~\etal~\cite{li2020docbank} presents a benchmark dataset named DocBank that contains 500K document pages with fine-grained token level annotations for DLA.
Some researchers solve this challenge by using the method of generating samples.
Li~\etal~\cite{li2020layoutgan}~proposed a method named LayoutGAN that synthesizes layouts samples by modeling geometric relations of different types of 2D elements.
Yang~\etal~\cite{yang2017learning}~used LATEX as the engine to synthesize data, control the synthesis parameters through programs to generated many examples with annotations.
However, the previous methods mostly focused on synthesizing the Manhattan layout examples, ignoring the non-Manhattan layout examples that exist in the current document.

\noindent\textbf{Data augmentation.}
Data enhancement is an effective means to solve data limitations by expanding the data space.
The image augmentation algorithms include mixing images, random erasing, generative adversarial networks, geometric transformations, color space augmentations, kernel filters, feature space augmentation, adversarial training, neural style transfer, and meta-learning~\cite{shorten2019survey}.
To facilitate the training of deep learning models from limited source data, Salehinejad~\etal~\cite{ghiasi2020simple}~proposed a sampling method based on the radical transformation in the polar coordinate system for image enhancement.
Through systematic research, Ghiasi~\etal~\cite{salehinejad2018image}~found that the simple mechanism of randomly pasting objects is good enough and can provide reliable benefits on top of a strong baseline. In addition, a copy-paste and semi-supervised method are proposed to utilize additional data through pseudo-labels.
Following these excellent works, we employ the image layer modeling~\cite{chen2020flexible} for modeling to obtain a more realistic non-Manhattan document layout.

\section{Method}
\label{sec:method}

\begin{figure}[t]
  \centering
  \includegraphics[width=0.92\linewidth]{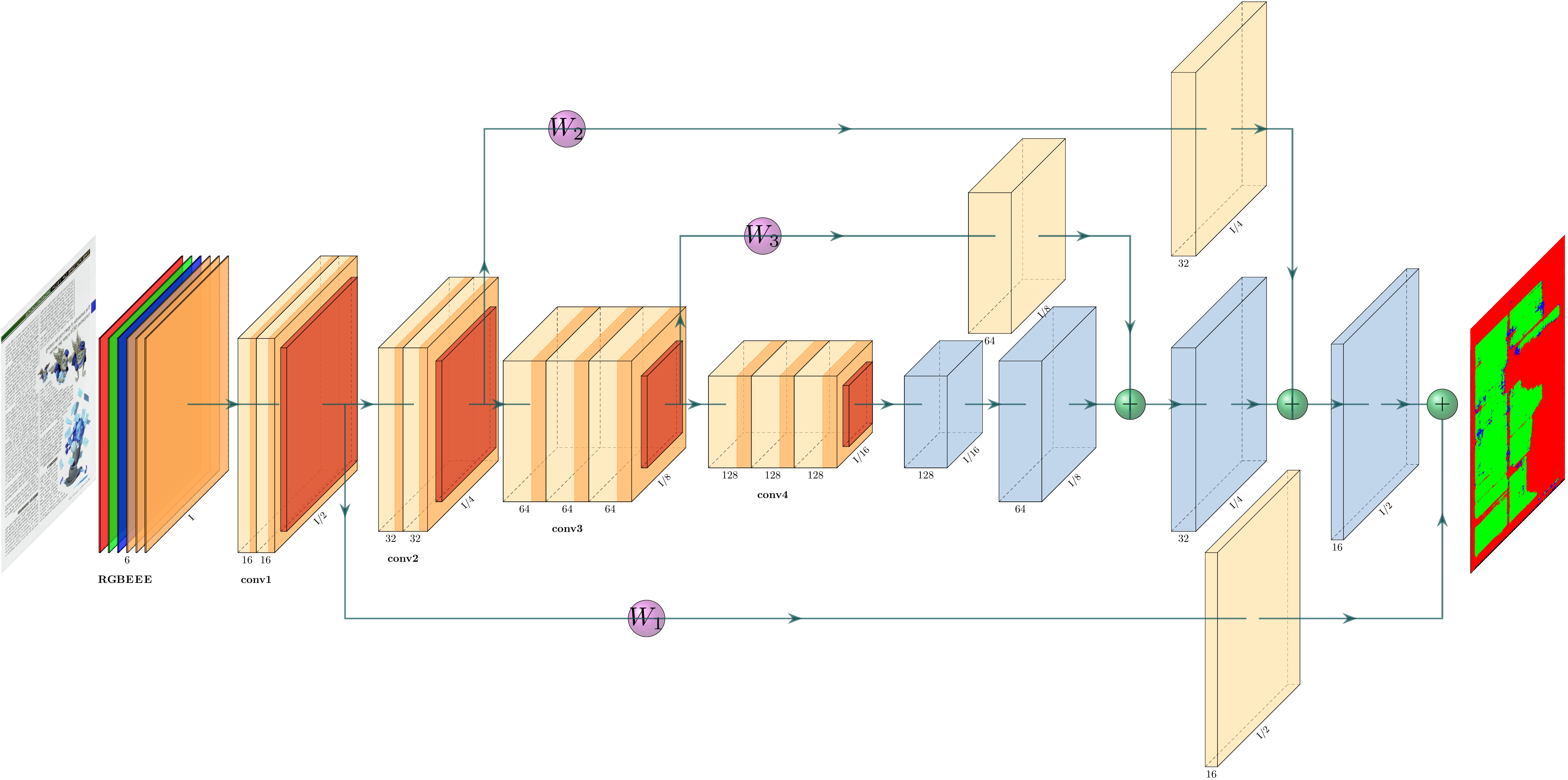}
  \caption{Architecture of the Lightweight Explicit Edge Embedding Network ($L\text{-}E^3Net$).}
  \label{fig:framework}
   \vspace{-8px}
  \end{figure}

In this section, we will introduce our framework and the details of the image layer modeling based on aesthetic guidance.

\subsection{Framework}
Following~\cite{wu2021ins}, in the document layout analysis task, the edge of the image will provide richer information, especially for the non-Manhattan layout documents.
The non-Manhattan layout document has more flexible and changeable edges, which can embedding richer information than the Manhattan layout document.
Following the work of $E^3Net$~\cite{wu2021ins}, we designed a more lightweight model named $L\text{-}E^3Net$ for processing non-Manhattan layout document, the parameter of the $L\text{-}E^3Net$ is only 0.7 M.
The architecture of the $L\text{-}E^3Net$ is shown in Fig.~\ref{fig:framework}.

\subsection{Image layer modeling}

At present, most of the work focuses on the Manhattan layout document, paid less attention to the non-Manhattan layout document.
However, in the real world, dealing with the Manhattan layout document is a challenge.
There are currently fewer documents with a non-Manhattan layout, so data expansion is the most effective solution.
However, the cost of data labeling for non-Manhattan layouts is expensive.
In this regard, we designed a data generation method based on the image layer inspired by Aesthetic-Guided and copy-paste.

\begin{figure}[t]
  \centering
  \includegraphics[width=0.88\linewidth]{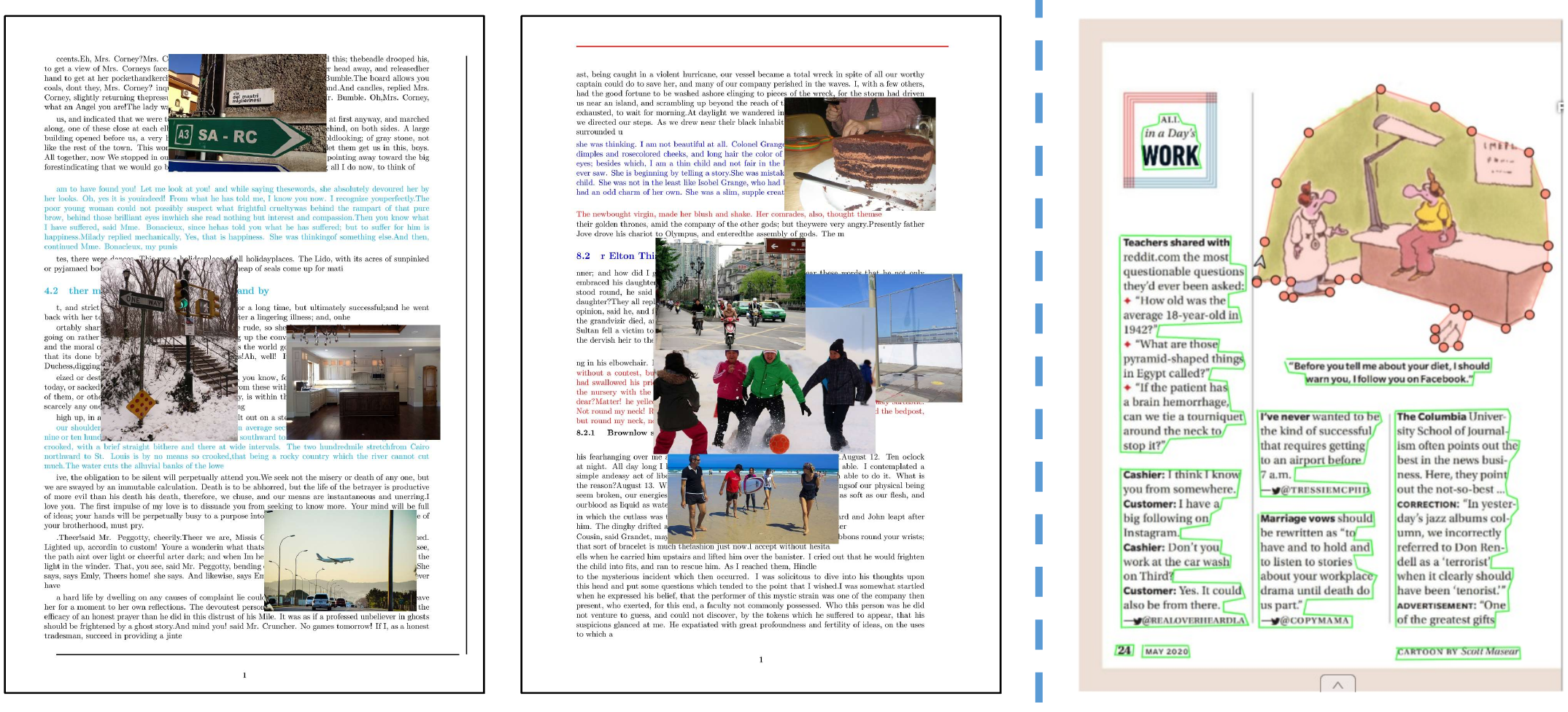}
  \caption{The example of synthesis by image layer modeling.}
  \label{fig:synthesis}
   \vspace{-8px}
  \end{figure}

As shown in Fig.~\ref{fig:synthesis}-Left, unlike the previous method, we no longer stitch text, tables, and images by the puzzle.
After the image is stored, it can only be stored in a rectangular manner, so direct stitching the image can only obtain the Manhattan layout document.
To tackle this challenge, we used the concept of the image layer.
As shown in Fig.~\ref{fig:motivation}, we first paste the text on the blank layout, and then cycle to select the appropriate position to paste the picture (supervise with some concepts of aesthetics), and finally get the layout effect shown in Fig.~\ref{fig:synthesis}-Left.
Specifically, our aesthetic guidance includes the following tasks. First, we ensure the relative size of the image during the process of pasting the image. We scale it according to the ratio, and the image size scaling factor is randomly selected within the interval [0.6, 1]; secondly, to ensure image layering, we randomly select image number is (0,8] and superimposition these images; finally,  to ensure the consistency of superimposed images, we measure the similarity of images during the image superimposition process, and the measurement process is represented by Equ.~\ref{equ:IIM}.

Suppose there are two image s ($m_s \times n_s$) and g ($m_g \times n_g$.), we use $\tau _{m,n}$ to represent the value of the image on $m, n$.
We measure the similarity of images by Equ.~\ref{equ:IIM}.

\begin{equation}
\begin{array}{l}
f(s,g) = \frac{1}{{256}}\sum\limits_{i = 0}^{255} {\left( {1 - \frac{{\left| {{s_i} - {g_i}} \right|}}{{Max\left( {{s_i},{g_i}} \right)}}} \right)} \\
{s_i} = \frac{{\sum {\left[ {{\tau _{{m_s},{n_s}}} = i} \right]} }}{{{m_s} \times {n_s}}},{g_i} = \frac{{\sum {\left[ {{\tau _{{m_g},{n_g}}} = i} \right]} }}{{{m_g} \times {n_g}}}
\end{array}
\label{equ:IIM}
\end{equation}

\subsection{Fine-grained annotations}
To measure the effectiveness of our method, we re-annotated a fine-grained page semantic structure extraction dataset (FPD).
We use the CVAT\footnote{https://github.com/openvinotoolkit/cvat} labeling platform for data labeling.
To obtain fine-grained labeling information, we use point labeling in the labeling process.
As shown in Fig.~\ref{fig:synthesis}-Right, it is very different from the rectangle or polygon labeling that is usually used at present.
The use of point labeling can mark the target object more prominently. To ensure the quality of labeling, we used a cross-check and review mechanism, we exported the label format to ``cvat for images 1.1 formats "~\cite{mouheb2021trodo}.
We will publish all datasets,  it includes the train data by using image layer modeling and the test data (FPD).

\section{Experiments}
\label{sec:exp}

In this section, we will introduce the experimental setup and evaluate the performance of the proposed framework.

\subsection{Experiment setup}

\vspace{0.08in}
\noindent\textbf{Datasets}.
\emph{FPD}~ is the first manual annotated dataset that focuses on fine-grained segmentation.
FPD consists of 66 non-Manhattan layout layouts document, which is manually annotated. In the labeling process, we use point labeling and polygon labeling, and the label export style is ``cvat for images 1.1". The data comes from the pages of complex Chinese and English magazines, and the page size is not fixed.

\emph{DSSE-200}~\cite{yang2017learning} is a comprehensive dataset that includes various dataset styles.
It contains 200 images, including pictures, PPT, brochure documents, old newspapers, and scan files with light changes.

\vspace{0.08in}
\noindent\textbf{Implementation details}.
Our model is implemented using PyTorch~\cite{paszke2019pytorch}.
We use the synthetic dataset to train the model, and use the FPD and DSSE-200 datasets as testset.
In order to verify the effectiveness of our method, we synthesized some data using the LATEX-based synthesis method.
Specifically, we synthesized 6000 images using the LATEX-based synthesis method~\cite{wu2021ins}, and synthesized 6000 images using image layer modeling.
We use $L_{D}$  to represent the network, we use $D$ represents the dataset used for training, $M$ represents the use of 3000 images that generated by LATEX and 3000 images generated by image layer modeling, $L$ represents the use of 6000 images synthesized by LATEX, and $I$ represents the use of 6000 images synthesized by image layer modeling.

We use the cross-entropy loss function (Eq.~\ref{equ:NLL}) train model and use Adam as the optimizer, we set the learning rate to $1e^{-3}$.

\begin{equation}
\begin{aligned}
  \mathcal{L}(x,label) &=  - {w_{label}}\log \frac{{{e^{{x_{label}}}}}}{{\sum\nolimits_{j = 1}^N {{e^{{x_j}}}} }}\\
   &= {w_{label}}\left( { - {x_{label}} + \log \sum\nolimits_{j = 1}^N {{e^{{x_j}}}} } \right),
\end{aligned}
\label{equ:NLL}
\end{equation}
where $x \in {\mathbb{R}^N}$ is the activation value without softmax, $N$ is the feature dimension of $x$, and the $label \in [0,C - 1]$ is the scalar of the corresponding label.
 $C$ is the number of classifications to be classified. $w \in {\mathbb{R}^C}$ represents the label weight.

\subsection{Results and discussion}

\begin{figure}[t]
\centering
\includegraphics[width=0.92\linewidth]{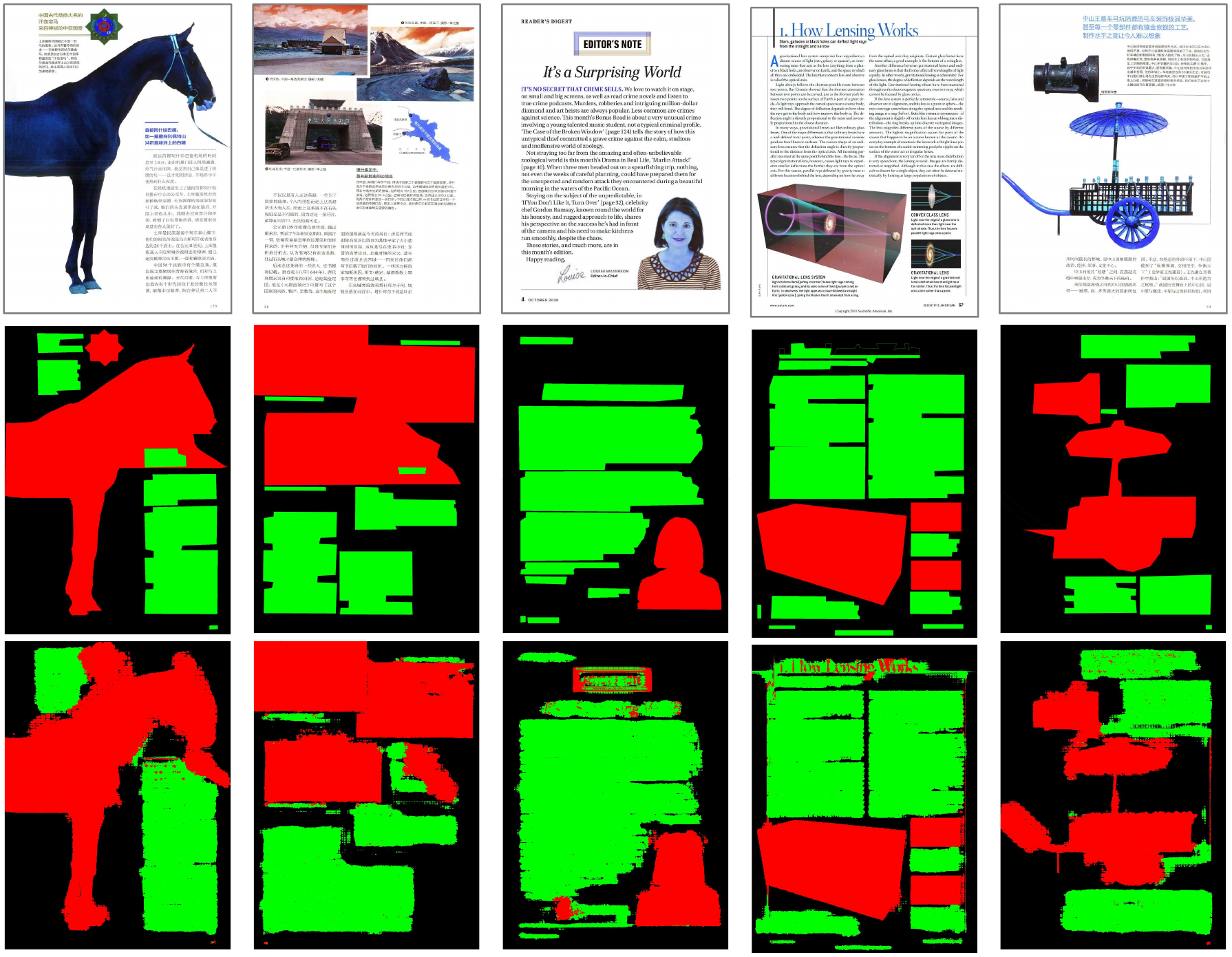}
\caption{Predicted result of FPD,
the first row is the original image, the second row is ground-true, and the third row is the predicted image.
Segmentation label colors are: \colorbox{red}{figure}, \colorbox{blue}{\color{white}{table}}, \colorbox{green}{text}, \colorbox{black}{\color{white}{background}}.}
\label{fig:CL}
\end{figure}

\vspace{0.08in}
\noindent\textbf{FPD}.
We also conducted experiments on FPD, and the results are shown in Table~\ref{tab:Result}. From the results in Table~\ref{tab:Result}, it can be seen that the use of layer-based sample synthesis can bring a significant improvement (67\% vs. 70\%).  We show the visualization results as shown in Fig.~\ref{fig:CL}. It can be seen that image layer modeling has a significant effect on determining the non-Manhattan layout area.

\vspace{0.08in}
\noindent\textbf{Comparison with prior arts}.
To verify the model and evaluate the layer modeling methods, we reproduced some classic methods and compared them with $L\text{-}E^3Net$.
We show the results in Table~\ref{tabel3}.
As can be seen from Table~\ref{tabel3}, the $L\text{-}E^3Net$ model has only 0.7M parameters, but it performed a comparable result.

\begin{table}[t]
\centering
\caption{Per-category comparison on both dataset.}
\begin{tabular}{p{20px}|p{12px}<{\centering}p{12px}<{\centering}p{12px}<{\centering}p{13px}<{\centering}||p{12px}<{\centering}p{12px}<{\centering}p{12px}<{\centering}p{12px}<{\centering}}
\toprule
\multirow{2}{*}{Model} & \multicolumn{4}{c||}{DSSE-200} & \multicolumn{4}{c}{FPD}  \\ \cline{2-5}\cline{6-9}
                       & Acc   & P     & R     & F1   & Acc  & P    & R    & F1   \\ \midrule
$L_{L}$      & \textbf{0.81}  & \textbf{0.84}  & \textbf{0.76}  & \textbf{0.79}& 0.78 & 0.66 & 0.68 & 0.67 \\
$L_{M}$                & 0.79  & 0.76  & 0.75  & 0.76 & 0.75 & 0.65 & 0.72 & 0.68 \\
$L_{I}$       & 0.79  & 0.76  & 0.75  & 0.75  & \textbf{0.83} & \textbf{0.67} & \textbf{0.74} & \textbf{0.70} \\ \bottomrule
\end{tabular}
\label{tab:Result}
\vspace{-7px}
\end{table}

\begin{figure}[t]
  \centering
  \includegraphics[width=0.92\linewidth]{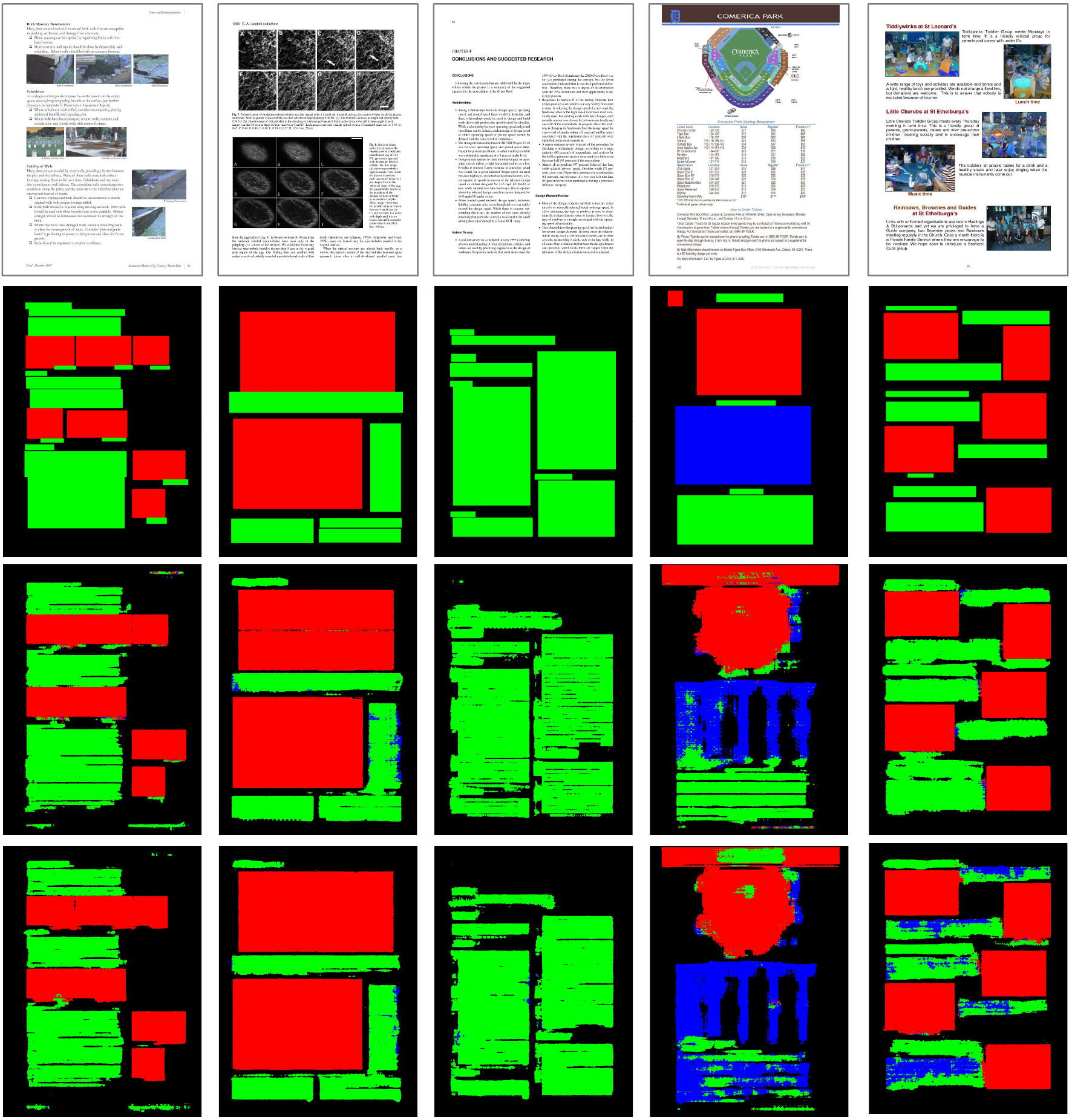}
  \caption{Predicted result of DSSE-200, the first row is the original image, the second row is ground-true, the third row is the prediction result of training using the mixed data set, and the fourth row is the prediction result of training using the LATEX synthesized dataset.
  }
  \label{fig:DSSE}
  \vspace{-8px}
  \end{figure}

 \vspace{0.08in}
\noindent\textbf{DSSE-200}.
We conducted experiments on DSSE-200, and the results are shown in Table~\ref{tab:Result}.
From the results in Table~\ref{tab:Result}, it can be seen that the effect of the model trained with the image layer modeling dataset is not as good as the model trained with only the LATEX synthetic dataset.
We visualized the results and showed them in Fig.~\ref{fig:DSSE}.
From the visualized results, it can be seen that the reason for this result is because DSSE is a dataset in Manhattan layout labeling, and its ground-true is based on regional division.
But using image layer modeling data will focus on fine-grained classification.
We evaluate based on pixels, the result will be worse after calculation than use the LATEX synthetic dataset.
However, it can be seen from Fig.~\ref{fig:DSSE} that the granularity of the divided regions of the model trained using the hybrid method is more significant for the detailed information.

\begin{table}[t]
	\centering
	\caption{The result on the FPD dataset(\%). The DV3+ means the DeeplabV3+ model with Xception backbone. The Para represents the trainable parameter amount of the model.}
	\begin{tabular}{lccccc}
		\toprule
		Method                                       &A                &P            &R             &F1     &Para \\ \midrule		
		Segnet~\cite{badrinarayanan2017segnet}       &82.5             &66.9         &74.1          &70.3   &29M  \\		
		PSPnet~\cite{zhao2017pyramid}                &83.9             &62.9         &72.1          &67.2   &46M  \\		
		PANet~\cite{li2018pyramid}                   &81.6             &68.2         &75.2          &71.5   &168M \\		
		DV3+~\cite{chen2018encoder}                  &79.9             &70.1         &69.4          &69.7   &53M  \\		
		$L\text{-}E^3Net$                            &83.1             &67.3         &73.7          &70.3   &0.7M  \\ \bottomrule

	\end{tabular}
	\label{tabel3}
\vspace{-6px}
\end{table}

\subsection{Ablation study}

\vspace{0.08in}
\noindent\textbf{Layer-based modeling.}
It is not difficult to see from Table~\ref{tab:Result} that the use of layer modeling can better handle non-Manhattan layout examples. It is worth pointing out that our generation method is under the guidance of aesthetics. We use non-aesthetic guidance to generate the dataset, and found that the result is not as good as the data set using aesthetics guidance (65\% vs. 70\%).

\vspace{0.08in}
\noindent\textbf{Model architecture.}
As we showed in Table~\ref{tabel3}, when the data reaches a certain level, the performance gap between the frameworks is not particularly obvious, so we chose a lightweight $L\text{-}E^ 3Net$. From the processing results, due to the consideration of edge information, the detailed processing of $L\text{-}E^3Net$ is also somewhat competitive.

\section{Conclusions}
\label{sec:conclusion}

In this paper, we proposed an unsupervised layer-based data synthesis method to tackle the document layout analysis task of non-Manhattan layouts, which is carried out under the guidance of aesthetic design principles.
To better capture the information of non-Manhattan layout documents, we propose a lightweight edge embedding network.
To better measure the proposed methods and models, this paper proposes a hand-labeled fine-grained document layout analysis data set for the first time.
The experimental results show that our method can better help the model to analyze the document layout.

\small{
\bibliographystyle{IEEEtran}
\bibliography{total}}

\begin{thebibliography}{10}
\providecommand{\url}[1]{#1}
\csname url@samestyle\endcsname
\providecommand{\newblock}{\relax}
\providecommand{\bibinfo}[2]{#2}
\providecommand{\BIBentrySTDinterwordspacing}{\spaceskip=0pt\relax}
\providecommand{\BIBentryALTinterwordstretchfactor}{4}
\providecommand{\BIBentryALTinterwordspacing}{\spaceskip=\fontdimen2\font plus
\BIBentryALTinterwordstretchfactor\fontdimen3\font minus
  \fontdimen4\font\relax}
\providecommand{\BIBforeignlanguage}[2]{{%
\expandafter\ifx\csname l@#1\endcsname\relax
\typeout{** WARNING: IEEEtran.bst: No hyphenation pattern has been}%
\typeout{** loaded for the language `#1'. Using the pattern for}%
\typeout{** the default language instead.}%
\else
\language=\csname l@#1\endcsname
\fi
#2}}
\providecommand{\BIBdecl}{\relax}
\BIBdecl

\bibitem{binmakhashen2019document}
G.~M. Binmakhashen and S.~A. Mahmoud, ``Document layout analysis: A
  comprehensive survey,'' \emph{ACM Computing Surveys}, vol.~52, no.~6, pp.
  1--36, 2019.

\bibitem{degtyarenko2021hierarchical}
I.~Degtyarenko, I.~Deriuga, A.~Grygoriev, S.~Polotskyi, V.~Melnyk,
  D.~Zakharchuk, and O.~Radyvonenko, ``Hierarchical recurrent neural network
  for handwritten strokes classification,'' in \emph{ICASSP}, 2021, pp.
  2865--2869.

\bibitem{li2021few}
Y.~Li, P.~Zhang, X.~Xu, Y.~Lai, F.~Shen, L.~Chen, and P.~Gao, ``Few-shot
  prototype alignment regularization network for document image layout
  segementation,'' \emph{Pattern Recognition}, vol. 115, p. 107882, 2021.

\bibitem{wu2021document}
X.~Wu, Z.~Hu, X.~Du, J.~Yang, and L.~He, ``Document layout analysis via dynamic
  residual feature fusion,'' in \emph{ICME}, 2021, pp. 1--6.

\bibitem{arroyo2021variational}
D.~M. Arroyo, J.~Postels, and F.~Tombari, ``Variational transformer networks
  for layout generation,'' in \emph{CVPR}, 2021, pp. 13\,642--13\,652.

\bibitem{sun2005page}
H.-M. Sun, ``Page segmentation for manhattan and non-manhattan layout documents
  via selective crla,'' in \emph{ICDAR}, 2005, pp. 116--120.

\bibitem{shen2020large}
Z.~Shen, K.~Zhang, and M.~Dell, ``A large dataset of historical japanese
  documents with complex layouts,'' in \emph{CVPR Workshops}, 2020, pp.
  548--549.

\bibitem{lombardi2020deep}
F.~Lombardi and S.~Marinai, ``Deep learning for historical document analysis
  and recognition: a survey,'' \emph{Journal of Imaging}, vol.~6, no.~10, p.
  110, 2020.

\bibitem{bhowmik2021binyas}
S.~Bhowmik, S.~Kundu, and R.~Sarkar, ``Binyas: a complex document layout
  analysis system,'' \emph{Multimedia Tools and Applications}, vol.~80, no.~6,
  pp. 8471--8504, 2021.

\bibitem{zhong2019publaynet}
X.~Zhong, J.~Tang, and A.~J. Yepes, ``Publaynet: largest dataset ever for
  document layout analysis,'' in \emph{ICDAR}, 2019, pp. 1015--1022.

\bibitem{li2020docbank}
M.~Li, Y.~Xu, L.~Cui, S.~Huang, F.~Wei, Z.~Li, and M.~Zhou, ``Docbank: A
  benchmark dataset for document layout analysis,'' in \emph{International
  Conference on Computational Linguistics}, 2020, pp. 949--960.

\bibitem{clark2016pdffigures}
C.~Clark and S.~Divvala, ``Pdffigures 2.0: Mining figures from research
  papers,'' in \emph{ACM/IEEE on Joint Conference on Digital Libraries}, 2016,
  pp. 143--152.

\bibitem{zhang2008active}
L.~Zhang, Y.~Tong, and Q.~Ji, ``Active image labeling and its application to
  facial action labeling,'' in \emph{ECCV}, 2008, pp. 706--719.

\bibitem{yang2017learning}
X.~Yang, E.~Yumer, P.~Asente, M.~Kraley, D.~Kifer, and C.~Lee~Giles, ``Learning
  to extract semantic structure from documents using multimodal fully
  convolutional neural networks,'' in \emph{CVPR}, 2017, pp. 5315--5324.

\bibitem{li2020layoutgan}
J.~Li, J.~Yang, A.~Hertzmann, J.~Zhang, and T.~Xu, ``Layoutgan: Synthesizing
  graphic layouts with vector-wireframe adversarial networks,'' \emph{IEEE
  Transaction on Pattern Analysis and Machine Intelligence}, 2020.

\bibitem{jyothi2019layoutvae}
A.~A. Jyothi, T.~Durand, J.~He, L.~Sigal, and G.~Mori, ``Layoutvae: Stochastic
  scene layout generation from a label set,'' in \emph{ICCV}, 2019, pp.
  9895--9904.

\bibitem{chen2020flexible}
J.~Chen, P.~Mu, R.~Liu, X.~Fan, and Z.~Luo, ``Flexible bilevel image layer
  modeling for robust deraining,'' in \emph{ICME}, 2020, pp. 1--6.

\bibitem{deng2017image}
Y.~Deng, C.~C. Loy, and X.~Tang, ``Image aesthetic assessment: An experimental
  survey,'' \emph{IEEE Signal Processing Magazine}, vol.~34, no.~4, pp.
  80--106, 2017.

\bibitem{salehinejad2018image}
H.~Salehinejad, S.~Valaee, T.~Dowdell, and J.~Barfett, ``Image augmentation
  using radial transform for training deep neural networks,'' in \emph{ICASSP},
  2018, pp. 3016--3020.

\bibitem{kise1998segmentation}
K.~Kise, A.~Sato, and M.~Iwata, ``Segmentation of page images using the area
  voronoi diagram,'' \emph{Computer Vision and Image Understanding}, vol.~70,
  no.~3, pp. 370--382, 1998.

\bibitem{eskenazi2015delaunay}
S.~Eskenazi, P.~Gomez-Kr{\"a}mer, and J.-M. Ogier, ``The delaunay document
  layout descriptor,'' in \emph{ACM Symposium on Document Engineering}, 2015,
  pp. 167--175.

\bibitem{wahl1982block}
F.~M. Wahl, K.~Y. Wong, and R.~G. Casey, ``Block segmentation and text
  extraction in mixed text/image documents,'' \emph{Computer Graphics and Image
  Processing}, vol.~20, no.~4, pp. 375--390, 1982.

\bibitem{shafait2010effect}
F.~Shafait and T.~M. Breuel, ``The effect of border noise on the performance of
  projection-based page segmentation methods,'' \emph{IEEE Transaction on
  Pattern Analysis and Machine Intelligence}, vol.~33, no.~4, pp. 846--851,
  2010.

\bibitem{shorten2019survey}
C.~Shorten and T.~M. Khoshgoftaar, ``A survey on image data augmentation for
  deep learning,'' \emph{Journal of Big Data}, vol.~6, no.~1, pp. 1--48, 2019.

\bibitem{ghiasi2020simple}
G.~Ghiasi, Y.~Cui, A.~Srinivas, R.~Qian, T.-Y. Lin, E.~D. Cubuk, Q.~V. Le, and
  B.~Zoph, ``Simple copy-paste is a strong data augmentation method for
  instance segmentation,'' in \emph{CVPR}, 2021, pp. 2918--2928.

\bibitem{wu2021ins}
X.~Wu, Y.~Zheng, T.~Ma, H.~Ye, and L.~He, ``Document image layout analysis via
  explicit edge embedding network,'' \emph{Information Sciences}, vol. 577, pp.
  436--448, 2021.

\bibitem{mouheb2021trodo}
K.~Mouheb, A.~Y{\"u}rekli, and B.~Y{\i}lmazel, ``Trodo: A public vehicle
  odometers dataset for computer vision,'' \emph{Data in Brief}, p. 107321,
  2021.

\bibitem{paszke2019pytorch}
A.~Paszke, S.~Gross, F.~Massa, A.~Lerer, J.~Bradbury, G.~Chanan, T.~Killeen,
  Z.~Lin, N.~Gimelshein, L.~Antiga, A.~Desmaison, A.~Kopf, E.~Yang, Z.~DeVito,
  M.~Raison, A.~Tejani, S.~Chilamkurthy, B.~Steiner, L.~Fang, J.~Bai, and
  S.~Chintala, ``Pytorch: An imperative style, high-performance deep learning
  library,'' in \emph{NeurIPS}, vol.~32, 2019, pp. 8026--8037.

\bibitem{badrinarayanan2017segnet}
V.~Badrinarayanan, A.~Kendall, and R.~Cipolla, ``Segnet: A deep convolutional
  encoder-decoder architecture for image segmentation,'' \emph{IEEE Transaction
  on Pattern Analysis and Machine Intelligence}, vol.~39, no.~12, pp.
  2481--2495, 2017.

\bibitem{zhao2017pyramid}
H.~Zhao, J.~Shi, X.~Qi, X.~Wang, and J.~Jia, ``Pyramid scene parsing network,''
  in \emph{CVPR}, 2017, pp. 2881--2890.

\bibitem{li2018pyramid}
H.~Li, P.~Xiong, J.~An, and L.~Wang, ``Pyramid attention network for semantic
  segmentation,'' \emph{BMVC}, 2018.

\bibitem{chen2018encoder}
L.-C. Chen, Y.~Zhu, G.~Papandreou, F.~Schroff, and H.~Adam, ``Encoder-decoder
  with atrous separable convolution for semantic image segmentation,'' in
  \emph{ECCV}, 2018, pp. 801--818.

\end{thebibliography}

\end{document}